\documentclass[conference]{IEEEtran}
\ifCLASSINFOpdf
   \usepackage[pdftex]{graphicx}
   \graphicspath{{../pdf/}{../jpg/}}
   \DeclareGraphicsExtensions{.pdf,.jpg,.png}
\else
   \usepackage[dvips]{graphicx}
   \graphicspath{{../eps/}}
   \DeclareGraphicsExtensions{.eps}
\fi
\begin{document}
%
\title{Facial Emotion Recognition using Convolutional Neural Networks}

\author{\IEEEauthorblockN{Akash Saravanan}
\IEEEauthorblockA{Department of Computer \\Science \& Engineering\\
Sri Venkateswara College \\of Engineering\\
Anna University, Chennai, \\Tamil Nadu, India\\
Email: akashsara@outlook.com}
\and
\IEEEauthorblockN{Gurudutt Perichetla}
\IEEEauthorblockA{Department of Computer \\Science \& Engineering\\
Sri Venkateswara College \\of Engineering\\
Anna University, Chennai, \\Tamil Nadu, India\\
Email: guruduttperichetla@gmail.com}
\and
\IEEEauthorblockN{Dr. K.S.Gayathri}
\IEEEauthorblockA{Associate Professor\\
Department of Computer \\Science \& Engineering\\
Sri Venkateswara College \\of Engineering\\
Anna University, Chennai, \\Tamil Nadu, India\\
Email: gayasuku@svce.ac.in}}

\maketitle

\begin{abstract}
Facial expression recognition is a topic of great interest in most fields from artificial intelligence and gaming to marketing and healthcare. The goal of this paper is to classify images of human faces into one of seven basic emotions. A number of different models were experimented with, including decision trees and neural networks before arriving at a final Convolutional Neural Network (CNN) model. CNNs work better for image recognition tasks since they are able to capture spacial features of the inputs due to their large number of filters. The proposed model consists of six convolutional layers, two max pooling layers and two fully connected layers. Upon tuning of the various hyperparameters, this model achieved a final accuracy of 0.60.
\end{abstract}


%
\IEEEpeerreviewmaketitle

\section{Introduction}
Human beings communicate with each other in the form of speech, gestures and emotions. As such systems that can recognize the same are in great demand in many fields. With respect to artificial intelligence, a computer will be able to interact with humans much more naturally if they are capable of understanding human emotion. It would also help during counseling and other health care related fields. In an E-Learning system, the presentation style may be varied depending on the student's state. However in many cases, static emotion detection is not very useful. It is essential to know the user's feelings over a period of time in a live environment. Thus, the paper proposes a model that is aimed at real-time facial emotion recognition.

For real-time purposes, facial emotion recognition has a number of applications. Facial emotion recognition could be used in conjunction with other systems to provide a form of safety. For instance, ATMs could be set up such that they won't dispense money when the user is scared. In the gaming industry, emotion-aware games can be developed which could vary the difficulty of a level depending on the player's emotions. It also has uses in video game testing. At present, players usually give some form of verbal or written feedback. Using facial emotion recognition, the number of testers can be increased to accomodate people who use different languages or people who are unable to cohesively state their opinions on the game. By judging their expressions during different points of the game, a general understanding of the game's strong and weak points can be discerned. Emotions can also be gauged while a viewer watches ads to see how they react to them. This is especially helpful since ads do not usually have feedback mechanisms apart from tracking whether the ad was watched and whether there was any user interaction. Software for cameras can use emotion recognition to take photos whenever a user smiles. 

However not all emotions can be inferred just by looking at someone. In his 1971 paper, Paul Ekman et al. \cite{Ekman} identified six basic, universal facial expressions - anger, disgust. fear, happiness, sadness and surprise. Even today, researchers aim to identify these six emotions with reliable accuracy. Emotions can be inferred from a person's actions, speech, writing and facial expressions. In terms of facial emotion recognition, one major challenge lies in the data collected. Most datasets contain labelled images which are generally posed. This generally involves photos taken in a stable environment such as a laboratory. While it is much easier to accurately predict the emotion in such scenarios, these systems tend to be unreliable in predicting emotions in the "wild" (Uncontrolled environments). Another issue is that most datasets are from these controlled environments and it is relatively harder to obtain labelled datasets of emotions in the wild. Furthermore, most datasets have relatively lesser training data for emotions such as fear and disgust when compared to emotions such as happiness. Another factor to take into account is a person's pose. It is significantly harder to determine the emotion of a person when only half of their face is visible. In addition, lighting plays a major role in facial emotion recognition. Systems may fail to identify an emotion that it normally would identify if the lighting conditions are poor. Finally, one must remember that a user's emotional state is a combination of many factors, a smile does not always mean that a person is genuinely happy. 

The objective of this project is to classify human faces into one of the six universal emotions or a seventh neutral emotion. In recent years, many papers have been published that use deep learning for facial emotion recognition \cite{Kahou} \cite{Vivek} \cite{Zhang}. These papers used freely available datasets with state of the art models \cite{Ali} achieving an accuracy of 0.66. With this in mind, a number of different models both new and old were experimented with to arrive at a final model with comparable results. 


\section{Related Work}

Yu and Zhang \cite{Zhang} used a five layer ensemble CNN to achieve a 0.612 accuracy. They pre-trained their models on the FER-2013 dataset and then finetuned the model on the Static Facial Expressions in the Wild 2.0 (SFEW) \cite{SFEW} dataset. They used an ensemble of three face detectors to detect and extract faces from the labelled movie frames of SFEW. They then proposed a data perturbation and voting method to increase the recognition performance of the CNN. They also chose to use stochastic pooling layers over max pooling layers citing its better performance on their limited data.

Kahou et al. \cite{Kahou} used a CNN-RNN architecture to train a model on individual frames of videos as well as static images. They made use of the Acted Facial Expressions in the Wild (AFEW) \cite{AFEW} 5.0 dataset for the video clips and a combination of the FER-2013 and Toronto Face Database for the images. Instead of using long short term memory (LSTM) units, they used IRNNs \cite{IRNN} which are composed of rectified linear units (ReLUs).  These IRNNs provided a simple mechanism for dealing with the vanishing and exploding gradient problem. They achieved an overall accuracy of 0.528.

Mollahosseini et al. \cite{Ali} proposed a network consisting of two convolutional layers each followed by max pooling and then four Inception layers. They used this network on seven different datasets including the FER-2013 dataset. They also compared the accuracies of their proposed network with an AlexNet \cite{AlexNet} network trained on the same datasets. They found that their architecture had better performance on the MMI and FER-2013 datasets with comparable performances on the remaining five datasets. The FER-2013 dataset in particular managed to reach an accuracy of 0.664.

Ming Li et al. \cite{Ming} propose a neural network model to overcome two shortcomings in still image based FERs which are the inter-variability of emotions between subjects and misclassification of emotions. The model consists of two convolutional neural networks - the first is trained with facial expression databases whereas the second is a DeepID network used for learning identity features. These two networks are then concatenated together as a Tandem Facial Expression of TFE Feature which is fed to the fully connected layers to form a new model. The proposed model was evaluated on two datasets, namely the FER+ database and the Extended Cohn-Kanade (CK+) database. The identity features were learned from the CASIA-WebFace database. The model was trained for 200 epochs and achieved an accuracy measure of 71.1\% on the FER2013 dataset, 99.31\% on the CK+ database. These experimental results show that the model outperforms many state-of-the-art methods on the CK+ and FER+ databases.

Tan et al. \cite{Tan} propose a neural network model to classify a group image into a particular emotion - positive, neutral or negative. The model consists of two convolutional neural networks - the first is based on group images and the second is based on individual facial emotions. The facial emotion CNN comprises of two CNNs - one for aligned faces which is trained using the ResNet64 model using the Webface dataset and the other for non-aligned faces which is trained using the ResNet34 model on the FER+ dataset. The group images are trained using VGG19 model on the Places and ImageNet datasets. Fine-tuning is done with batch normalisation and average pooling of the ResNet101 and BN-Inception models, with a dropout of 0.5. The validation set consists of 2068 images - combined from all of the datasets used for training and the model achieved an accuracy measure of 80.9.

Most other works in the same field attempted to solve the facial emotion recognition problem by the use of a combination of different datsets. In this paper, a single dataset, FER-2013 was chosen over such a combination of different datasets and then experiments were conducted with different models to find the highest accuracy that each model could reach. 

\section{Theoretical Background}
Three different types of models are used in this paper. This section details the theoretical background for each of these models.

\subsection{Decision Tree}
Decision trees are a supervised learning technique that predicts a value given a set of inputs by "learning" rules based on a set of training data. To put it simply, it is a massive tree of if-then-else rules. The decision making process starts off at the root of the tree and descends by answering a series of yes-no questions. At the end of this if-then-else chain, it arrives at a single predicted label. This is the output of a decision tree.

\subsection{Feedforward Neural Network}
A neural network is a system of algorithms that attempts to identify underlying relationships in a set of data by using a method that mimics the way in which a human brain operates. Neural networks consist of nodes connected to each other through edges. Each connection has a weight and a bias. A weight is the strength of the connection. The greater the weight, the greater impact it will have on the final output. A bias is a minimum threshhold which the sum of all the weighted inputs must cross. Neural networks are primarily employed for classification tasks. Neural networks consist of three layers - input, output and hidden layers. Hidden layers are sets of features based on the previous layer. They are intermediate layers in the network. A neural network recognises objects based on the concept of learning. Learning consists of six steps. Initially weights are initialised and a batch of data is fetched. This data is known as training data. A forward propagation is done on the data by passing through the network. A metric of difference between expected output and actual output is computed by the use of activation functions which perform computations on the data based on standard mathematical distributions such as Hyperbolic tangent and Sigmoid. This is known as cost. The goal is to minimize or reduce the cost. For this purpose, gradients of cost and weight are backpropagated to know how to adjust the weights to reduce the cost. Backpropagation refers to a backward pass of the network. Later, the weights are updated and the whole process is repeated. Feed-forward networks are also termed as multi-layer perceptrons.

\subsection{Convolutional Neural Network}
A Convolutional neural network is a neural network comprised of convolution layers which does computational heavy lifting by performing convolution. Convolution is a mathematical operation on two functions to produce a third function. It is to be noted that the image is not represented as pixels, but as numbers representing the pixel value. In terms of what the computer sees, there will simply just be a matrix of numbers. The convolution operation takes place on these numbers. We utilize both fully-connected layers as well as convolutional layers. In a fully-connected layer, every node is connected to every other neuron. They are the layers used in standard feedforward neural networks. Unlike the fully-connected layers, convolutional layers are not connected to every neuron. Connections are made across localized regions. A sliding "window" is moved across the image. The size of this window is known as the kernel or the filter. They help recognise patterns in the data. For each filter, there are two main properties to consider - padding and stride. Stride represents the step of the convolution operation, that is, the number of pixels the window moves across. Padding is the addition of null pixels to increase the size of an image. Null pixels here refers to pixels with value of 0. If we have a 5x5 image and a window with a 3x3 filter, a stride of 1 and no padding, the output of the convolutional layer will be a 3x3 image. This condensation of a feature map is known as pooling. In this case, "max pooling" is utilized. Here, the maximum value is taken from each sliding window and is placed in the output matrix.

Convolution is very effective in image recognition and classification compared to a feed-forward neural network. This is because convolution allows to reduce the number of parameters in a network and take advantage of spatial locality. Further, convolutional neural networks introduce the concept of pooling to reduce the number of parameters by downsampling. Applications of Convolutional neural networks include image recognition, self-driving cars and robotics. CNN is popularly used with videos, 2D images, spectrograms, Synthetic Aperture Radars. 

\section{Experimental Setup}
This section details the data used for training and testing, how the data was preprocessed, the various models that were used and an evaluation of each model. 

\begin{figure}[h]
\centering
\includegraphics[width=3.5in, keepaspectratio]{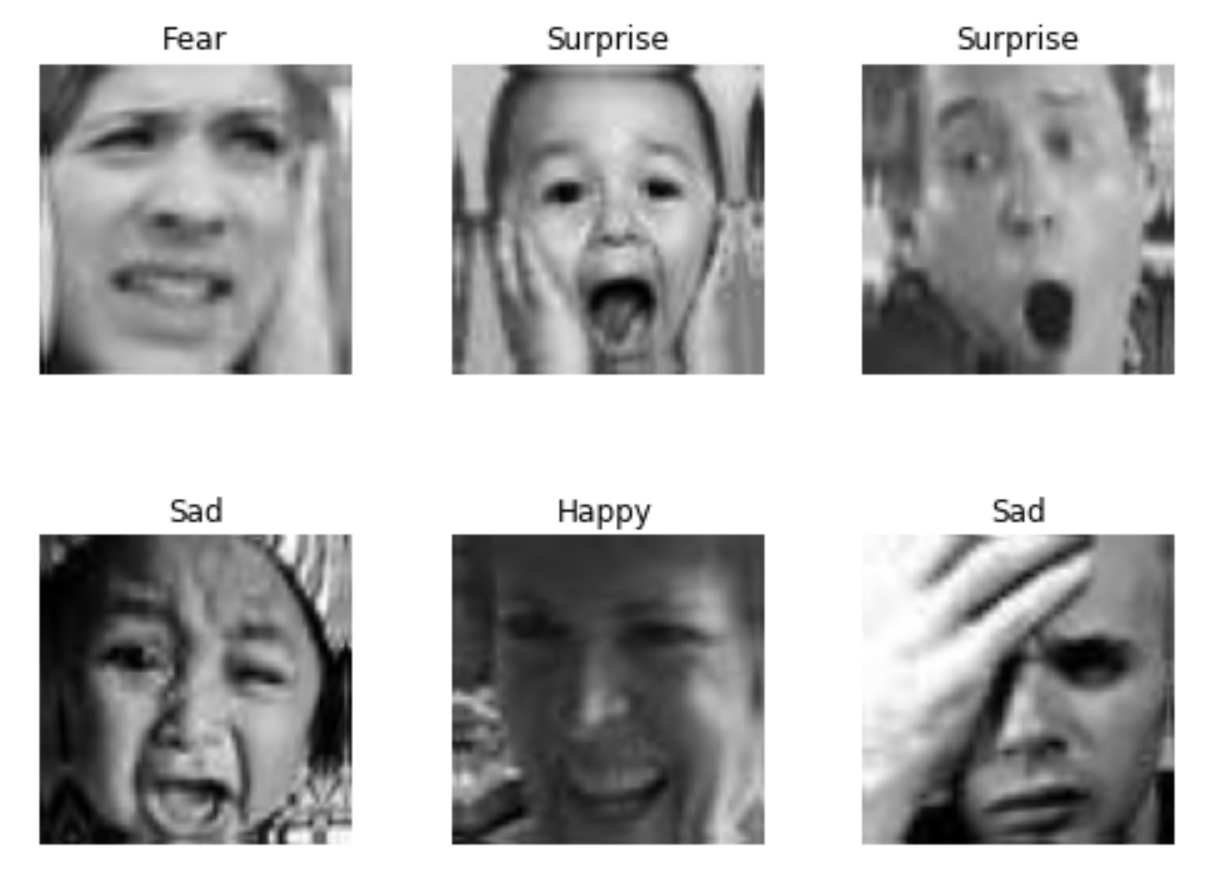}
\caption{FER-2013 Expressions}
\label{fig1}
\end{figure}

\subsection{Dataset}
In general, neural networks, especially deep neural networks, tend to perform better when larger amounts of training data set is present. With this in mind, the more popular Extended Cohn-Kanade (CK+) \cite{CKPlus} and the Static Facial Expressions in the Wild (SFEW) \cite{SFEW} datasets were overlooked. 

\begin{figure}[h]
\centering
\includegraphics[width=3.5in]{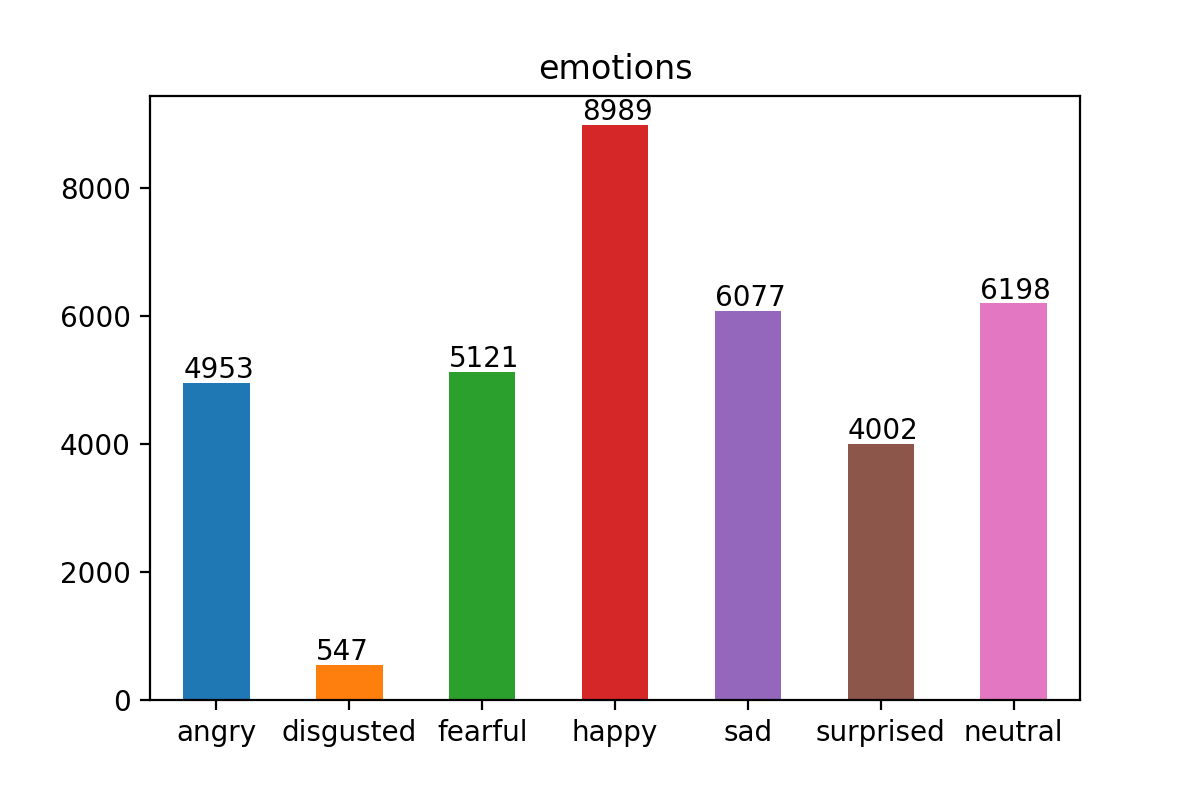}
\caption{FER-2013 Expression Distribution}
\label{fig2}
\end{figure}

Instead the Facial Expression dataset (FER-2013) was chosen. The FER-2013 dataset was introduced in the ICML 2013 Challenges in Representation Learning \cite{FER2013}. It contains 35,887 images with the following basic expressions: angry, disgusted, fearful, happy, sad, surprised and neutral. Figure \ref{fig1} shows the distribution of each expression. Each image is a frontal view of a subject, taken from the wild and annotated to one of the seven expressions. A sample of these expressions are shown in Figure \ref{fig2}. It is to be noted that the number of disgusted expressions (547) is much lower in comparison to the other expressions. There was also an obvious bias towards happy expressions due to the sheer number of sample data present for the expression. 

\subsection{Preprocessing}
The dataset consisted of a number of images represented as strings of 2304 space separated numbers which were then converted to a 48*48 matrix. Each number represented a pixel value. The original data of 35,887 images was split into a training set of 28,709 images and a testing set of 7,178 images - an 80:20 split. Generally, when it comes to deep learning, data is the biggest factor. The bigger the training set, the better the output. If there is less training data, there is a lot more variance in the final outputs due to a smaller set to train on. Bearing that in mind, having a testing set of 20\% of the total images may be seen as excessive. However, to prevent overfitting it is necessary to have a sizable testing set as well. It is also to be noted that in general, there is a 60-20-20 split up for training, testing \& validation respectively. For instance, Mollahosseini et al. \cite{Ali} divided their 275k image dataset into 60\% for training, 20\% for testing and 20\% for validation. In this case, the validation set was forewent in favor of retraining the entire model every time the hyper-parameters were tuned. While this required more time and computational power, it provided a bigger training set in the end. A one-hot encoding scheme was used for the labels rather than classifiying emotions with numbers from 0-6. 
During the live testing, Haar Cascades \cite{Haar} were used to identify a face. This identified face was then taken as an image, converted to gray-scale and downscaled to a 48*48 image. Thus the image was converted to a format identical to that which was used to train the model.

\subsection{Choosing a Model}
4 different models were used in order to select the best foundation to work on - 3 neural networks and a decision tree. The neural networks were implemented using Keras \cite{Keras} with a TensorFlow backend running in Python. The decision tree was implemented with the help of Sci-Kit Learn \cite{SKLearn}. All the implementations are written in Python 3.6 and are wholy reproducible with freely available software. A comparison of the final accuracy of all the models can be seen in Table \ref{tab1}. 

Since there are 4 different models used, the individual training algorithm for each model is detailed with the model description itself. The testing algorithm for the same is described in subsection D.

\begin{table}[h]
\renewcommand{\arraystretch}{1.3}
\caption{Comparison of Models}
\label{tab1}
\centering
\begin{tabular}{|c|c|}
\hline
Model & Accuracy\\
\hline
Decision Tree & 30.84\%\\
\hline
Feed Forward NN & 17.38\%\\
\hline
Simple CNN & 24.72\%\\
\hline
Proposed CNN & 55.61\%\\
\hline
\end{tabular}
\end{table}

\subsubsection{Decision Tree}
A decision tree was chosen at first chosen due to a combination of it requiring little to no effort in preparation of data as well as its reputation for working well in almost all scenarios. The first attempt was to implement a decision tree using Sci-Kit Learn. A standard decision tree classifier was used and the parameters were tuned. However the only parameter that showed a noticeable difference while tuning was the emph{min\_samples\_split} parameter. With the parameter set to 40, the overall accuracy was only a mere 0.309. As state of the art models reached accuracies greater than 0.6, it was decided to try out a neural network instead.

\begin{table}[]
\renewcommand{\arraystretch}{1.3}
\caption{Architecture of the Feedforward Neural Network}
\label{tabANN}
\centering
\begin{tabular}{c}
\hline
\textbf{Feed-Forward Neural Network}\\
\hline
FULLY CONNECTED\\
RELU\\
DROPOUT\\
\hline
FULLY CONNECTED\\
RELU\\
DROPOUT\\
\hline
FULLY CONNECTED\\
SOFTMAX\\
\hline
\end{tabular}
\end{table}

\subsubsection{Feedforward Neural Network}
The next attempt was to try out the most basic form of an artificial neural network, a Feedforward Neural Network. Three layers were used - an input layer, a single hidden layer and an output layer. Each of these layers was a fully connected (dense) layer. The architecture of this model can be seen in Table \ref{tabANN}. A dropout of rate 0.2 was applied for the input and hidden layer in an attempt to prevent overfitting. The output layer uses a softmax activation function while the remaining layers use ReLU (Rectified Linear Units). However this model ended up predicting the same expression for every input - angry. Further tuning of the hyperparameters lead to no difference and so it was decided that a convolutional neural network might work better.

\begin{table}[h]
\renewcommand{\arraystretch}{1.3}
\caption{Architecture of the Simple CNN}
\label{tabCNN}
\centering
\begin{tabular}{c}	
\hline
\textbf{Simple Convolutional Neural Network}\\
\hline
CONV2D-32\\
RELU\\
\hline
CONV2D-64\\
RELU\\
MAXPOOL2D\\
DROPOUT\\
\hline
FLATTEN\\
\hline
FULLY CONNECTED\\
RELU\\
DROPOUT\\
\hline
FULLY CONNECTED\\
SOFTMAX\\
\hline
\end{tabular}
\end{table}

\subsubsection{Simple Convolutional Network}
Next, a basic convolutional network was tried out. This model's architecture can be seen in Table \ref{tabCNN}. This model consisted of two two-dimensional convolutional layers followed by a two-dimensional max pooling layer which was followed by two fully-connnected(Dense) layers. The output was flattened before entering the fully connected layers. Dropout was applied to the max pooling and to the first fully connected layer to reduce overfitting. This model while more complicated, ended up having the same issue as the feedforward neural network except instead of predicting angry, it predicted happy for all inputs. This makes sense as a quarter of the inputs are for the happy expression. In order to allow the model to actually learn, attempts were made to make the model deeper.

\begin{figure*}[ht]
\includegraphics[width=7in, keepaspectratio]{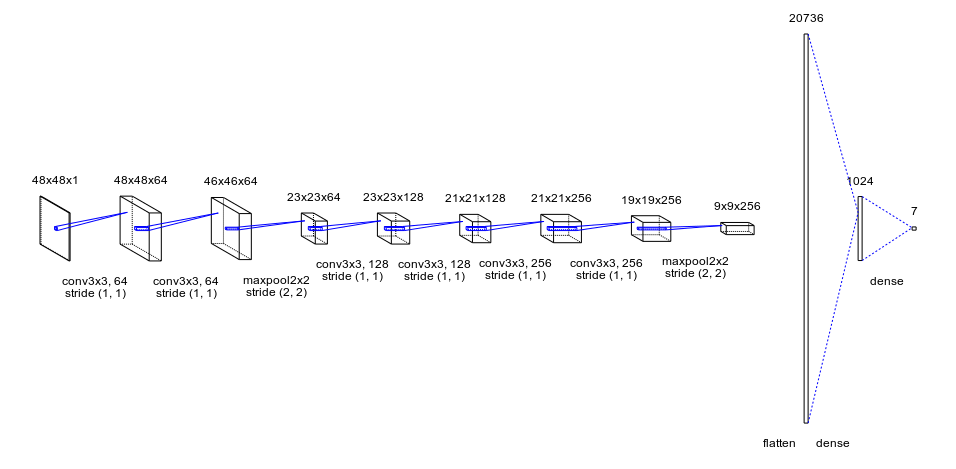}
\caption{Final Model Architecture}
\label{fig3}
\end{figure*}

\subsubsection{Final Model}
The final model is depicted in Table \ref{tabFinalCNN}. The network consists of six two-dimensional convolutional layers, two max pooling layers and two fully connected layers. Max pooling uses the maximum value from each of a cluster of neurons at the prior layer. This reduces the dimensionality of the output array. The input to the network is a preprocessed face of 48 x 48 pixels. The model was developed based on the observation of the performance of the previous models. It was decided to go with a deeper network over a wide one. The advantage of using more layers is that it prevents memorization. A wide but shallow network memorizes well but does not generalize well. Multi-layer networks learn features at levels of abstractions allowing them to generalize well. The number of layers were selected so as to maintain a high level of accuracy while still being fast enough for real-time purposes. The proposed CNN differs from a simple CNN in that it uses 4 more convolutional layers and each of its convolutional layers differ in filter size. In addition, it utilized max pooling and dropout more effectively in order to minimize overfitting. 

\begin{table}[h]
\renewcommand{\arraystretch}{1.3}
\caption{Architecture of the Proposed CNN}
\label{tabFinalCNN}
\centering
\begin{tabular}{c}
\hline
\textbf{Proposed Convolutional Neural Network}\\
\hline
CONV2D-64\\
RELU\\
\hline
CONV2D-64\\
RELU\\
MAXPOOL2D\\
DROPOUT\\
\hline
CONV2D-128\\
RELU\\
\hline
CONV2D-128\\
RELU\\
\hline
CONV2D-256\\
RELU\\
\hline
CONV2D-256\\
RELU\\
MAXPOOL2D\\
DROPOUT\\
\hline
FLATTEN\\
\hline
FULLY CONNECTED\\
RELU\\
DROPOUT\\
\hline
FULLY CONNECTED\\
SOFTMAX\\
\hline
\end{tabular}
\end{table}

The network consists of two convolutional layers with a filter size of 64 each. This is then followed by a max pooling layer. A dropout of rate 0.25 is applied to reduce overfitting. This is followed by a sequence of four convolutional layers. The first two have a filter size of 128 each and the latter two have a filter size of 256 each. A single max pooling layer follows these four layers with a dropout of rate 0.25. In order to convert the output into a single dimensional vector, the output of the previous layers was flattened. A fully connected layer with a L2 regularizer of penalty of 0.001 is then used alogn with an additional dropout of rate 0.5. Finally, a fully connected layer with a softmax activation function serves as the output layer. 

The kernel size, that is, the width and height of the 2D convolutional window is set to 3 x 3 for all convolutional layers. Each max pooling layer is two dimensional and uses a pool size of 2 x 2. This halves the size of the output after each pooling layer. All the layers bar the output layer used a ReLU activation function. The ReLU activation function is used here due to benefits such as sparsity and a reduced likelihood of vanishing gradient. The softmax activation function was used in the final output layer to receive the predicted probability of each emotion.

This model provided a base accuracy of 0.55 on the testing set. The hyperparamters were then tuned, namely the batch size, the optimizer and the number of epochs. Each model was set to run for 100 epochs. However, in the interest of saving time and computational power, the network was allowed to stop training if there was no change in the accuracy over consecutive epochs. That is, the network would stop training if there was no change in the accuracy over 4 continuous epochs. This saved both time and computational power, especially in cases where there was no change in the accuracy within the earlier epochs themselves. The decision turned out to be a good one as none of the models exceeded 20 epochs.

\subsection{Testing}
The dataset was initially split into an 80\%-training set and a 20\%-testing set. During the testing phase, each of the trained networks was loaded and fed the entire testing set one image at a time. This image was a new one which the model had never seen before. The image fed to the model was preprocessed in the same way as detailed in \ref{preprocessing}.  Thus the model did not know already what the correct output was and had to accurately predict it based on its own training. It attempted to classify the emotion shown on the image simply based on what it had already learned along with the characteristics of the image itself. Thus in the end, it gave a list of classified emotion probabilities for each image. The highest probability emotion for each image was then compared with the actual emotions associated with the images to count the number of accurate predictions.

The accuracy formula is detailed below. It simply counts the number of samples where the model correctly predicted the emotion and divides it by the total number of samples in the testing set. Here, the testing set consists of about 7,178 images.  
\begin{equation}
\label{accuracy}
Accuracy = \frac{Num. Correctly Predicted Emotions }{Total Num. Samples}
\end{equation}

\section{Results}
Upon tuning the hyperparameters, the highest accuracy was achieved for each optimizer. Using the RMSProp optimizer, an accuracy of 0.57 was reached over 20 epochs and a batch size of 96. The Stochastic Gradient Descent optimizer gave an accuracy of 0.55 out of the box and it could not be increased significantly by further tuning of the hyperparameters. Using the Adam optimizer with the default settings, a batch size of 64 and 10 epochs lead to an astoundingly low accuracy of 0.17. However upon setting the learning rate to 0.0001 and the decay to $10e-6$, the highest accuracy of 0.60 was attained. A comparison of the various hyperparameters that were tuned can be seen in Table \ref{tab2}.

\begin{table}[h]
\renewcommand{\arraystretch}{1.3}
\caption{Comparison of Hyperparameters}
\label{tab2}
\centering
\begin{tabular}{|c|c|c|c|}
\hline
Optimizer & Batch Size & Epochs & Accuracy\\
\hline
RMSProp & 64 & 24 & 55.96\%\\
\hline
RMSProp & 32 & 9 & 42.07\%\\
\hline
RMSProp & 96 & 20 & 57.39\%\\
\hline
SGD & 64 & 10 & 55.90\%\\
\hline
Adam & 64 &10 & 17.38\%\\
\hline
Adam & 128 & 20 & 60.58\%\\
\hline
\end{tabular}
\end{table}

Based on these results it can be concluded that the Adam optimizer which initially provided an abysmal accuracy turned out to be the best fit for the data. This makes sense as Adam is based off of RMSProp \& AdaGrad both of which are extensions of Stochastic Gradient Descent (SGD). It realizes the benefits of both RMSProp \& AdaGrad by utilizing an adaptive learning rate as well as bringing in momentum. In Adam (and RMSProp), the learning rate of each parameter is adaptively decided. Parameters that would ordinarily receive smaller or less frequent updates receive larger updates with Adam (the reverse is also true). This speeds up learning in cases where the appropriate learning rates vary across parameters. This is not the case with SGD which requires careful tuning of learning rates.

\begin{figure}[h]
\centering
\includegraphics[width=2.5in, keepaspectratio]{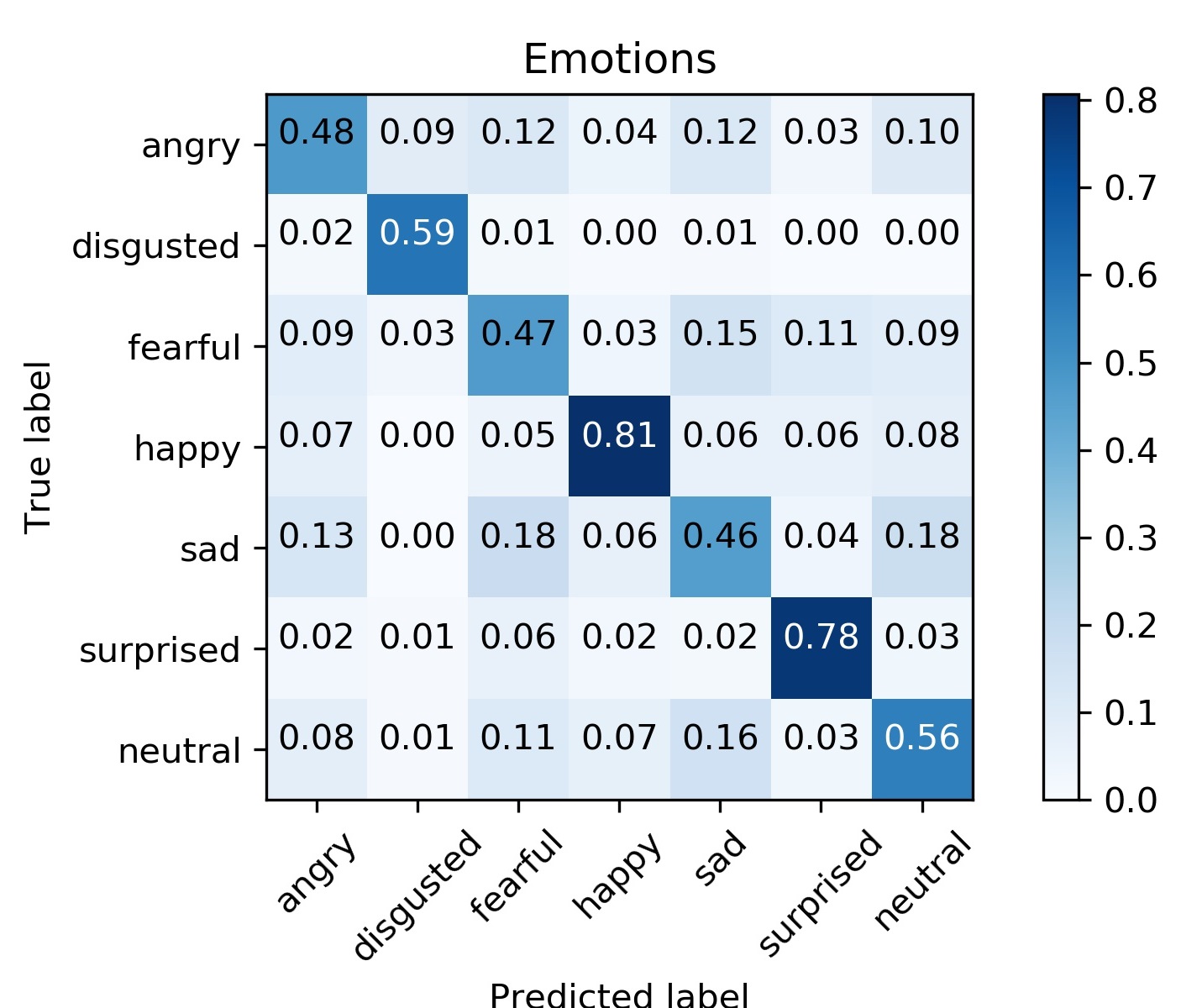}
\caption{Confusion Matrix for the final model}
\label{fig4}
\end{figure}

Adam and RMSProp differ only in that Adam uses the concept of Momentum. Like the physical phenomenon, momentum adds a fraction of the previous update to the current update, so that repeated updates in a particular direction compound. Thus a “momentum” is created, causing it to move faster and faster in that direction. Local optima are thus skipped due to the momentum and convergence is also sped up. It is also to be noted that the learning rate and decay had to be adjusted to arrive at a good accuracy. The reason why the Adam optimizer was initially unable to offer good results with the default learning rate of 0.001 and a decay of 0 was that it failed to converge. By setting the learning rate to a much smaller value in 0.0001 and the decay to $10e-6$, convergence actually occured. The decay value was set in such a way that over time the learning rate would further reduce. However accuracy alone doesn't paint the whole picture. A confusion matrix for the final model (using the Adam optimizer) was generated to take a look at how each individual emotion is dealt with. This matrix can be seen in Figure \ref{fig4}.

Understandably happiness is very easy to determine as a direct result of the number of sample data present. Interestingly the emotion of surprise reached nearly the same accuracy. The other emotions had lower but similar accuracies. Another point of interest is that it manages to determine the emotion of disgust a little more than half the time. The model when given an image (or a frame from a video) to predict from, does not simply give one final prediction. Rather it predicts a list of probabilities of each individual emotion. We then take the emotion with the highest probability as the final prediction. Thus we classify the status of the facial reaction based on the most probable emotion predicted by the model. Considering the sparse number of sample data, it is possible that the model may have been overfit.

To put this theory to the test, a small tool was developed that took a webcam's feed, detected faces, processed the faces and then fed it to the model. It was found that the model managed to predict almost all instances of happiness and most instances of surprise. It correctly predicted sadness and neutrality about half the time but it rarely predicted the other emotions correctly. The emotions of anger and fear in particular tended to mix while disgust was almost never predicted. It is also to be noted that in most cases of a wrong prediction, the second most likely prediction was often the right one. This indicates that the top-2 predicted emotions will be much more accurate. This is confirmed by Mollahosseini et al. \cite{Ali} Finally, during the live testing, it was observed that the model was able to predict emotions upon detecting a face instantaneously with no delays. 

\begin{figure}[h]
\centering
\includegraphics[width=2.5in, keepaspectratio]{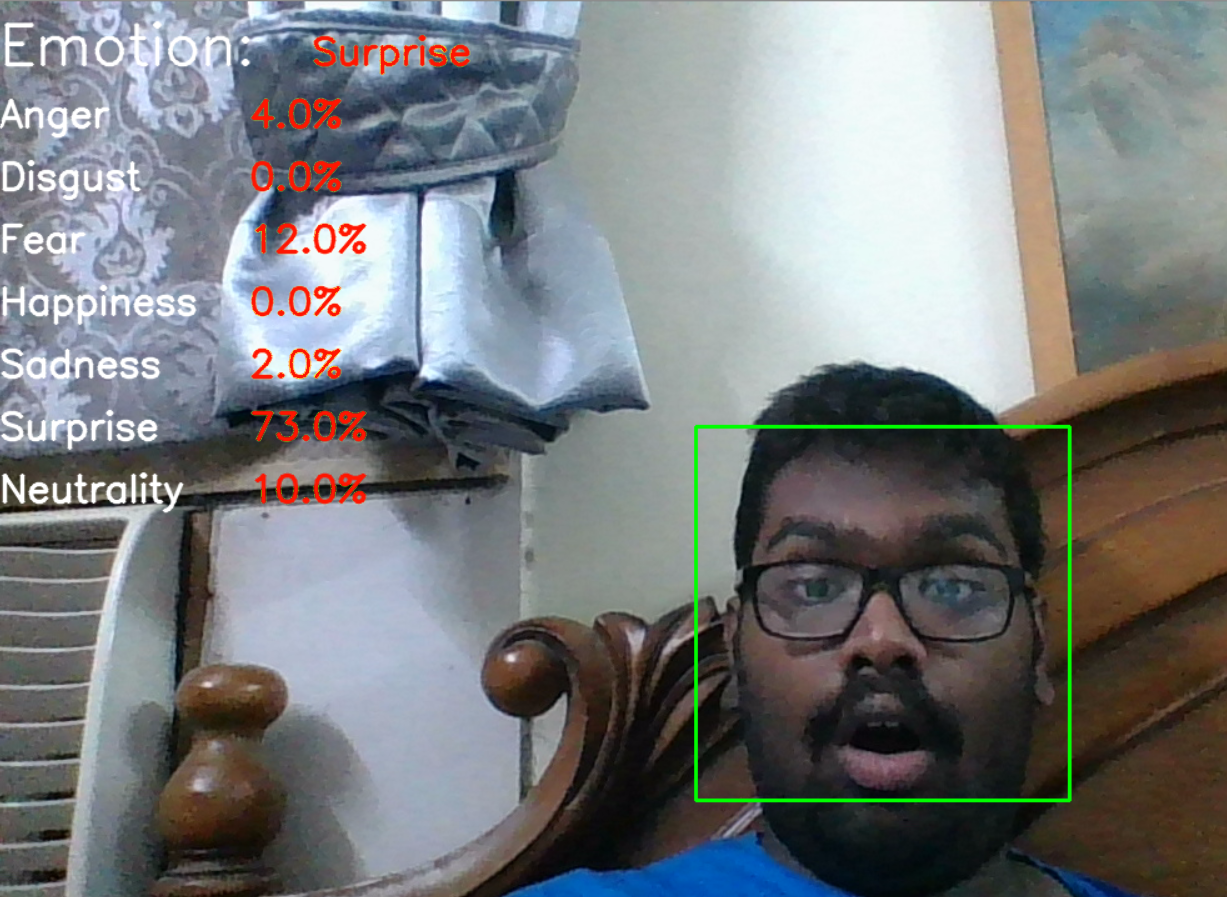}
\caption{Live Testing Module}
\label{fig5}
\end{figure}

\section{Conclusion \& Future Work}
In this paper, the aim was to classify facial expressions into one of seven emotions by using various models on the FER-2013 dataset. Models that were experimented with include decision trees, feed forward neural networks and smaller convolutional networks before arriving at the proposed model. The effects of different hyperparameters on the final model was then investigated. The final accuracy of 0.60 was achieved using the Adam optimizer with modified hyperparameters. It should also be noted that a nearly state-of-the-art accuracy was achieved with the use of a single dataset as opposed to a combination of many datasets. While it is true that other related works have managed to obtain higher accuracies - Mollahosseini et al.(0.66) and Yu and Zhang (0.61), they have used a combination of different datasets and large models in order to increase their overall accuracy. \ref{tab3} shows a comparison between the proposed approach and existing methodologies.

Given that only the FER-2013 dataset was used in this case without the use of other datasets, an accuracy of 0.60 is admirable as it demonstrates the efficiency of the model. In other words, the model demonstrated has used significantly less data for training and a deep but simple architecture to attain near-state-of-the-art results. 

At the same time, it also has its shortcomings. While the model did attain near-state-of-the-art results, it also means that it did not achieve state-of-the-art. Additionally, the relatively lower amount of data for emotions such as "disgust" make the model have difficulty predicting it. This however does illuminate a path for future work. If provided with more training data while still retaining the same network structure, the efficiency of the proposed system will be enhanced considerably. Sang, Dat and Thuan \cite{Sang} who used the same dataset, augmented the data to greatly increase the size of the training set to achieve similar results. Thus augmenting the existing data to enlarge the dataset might also prove to be a worthwhile avenue to explore. 

The ability of the model to make predictions in effectively real-time, indicates that real world uses of facial emotion recognition is barred only by the relative inaccuracies of the model itself. In the future, an indepth analysis of the top-2 predicted emotions may lead to a much more accurate and reliable system. Further training samples for the more difficult to predict emotion of disgust will definitely be required in order to perfect such a system. 

The real-time capacity of the model in addition to its quick training time and near-state-of-the-art accuracy allows the model to be adapted and used in nearly any use-case. This also implies that with some work, the model could very well be deployed into real-life applications for effective utilization in domains such as in healthcare, marketing and the video game industry.







\begin{thebibliography}{1}
\bibitem{Ekman}
P. Ekman, \& W. V. Friesen. Constants across cultures in the face and emotion. \emph{Journal of Personality and Social Psychology}, 17(2), 124-129. (1971)

\bibitem{FER2013}
Challenges in representation learning: Facial expression recognition challenge. (2013) http://www.kaggle.com/c/challenges-in-representation-learning-facial-expression-recognition-challenge

\bibitem{Zhang}
Zhiding Yu \& Cha Zhang. (2015). Image based Static Facial Expression Recognition with Multiple Deep Network Learning. \emph{Proceedings of the 2015 ACM on International Conference on Multimodal Interaction }. 435-442. 

\bibitem{Vivek}
Raghuvanshi, A., \& Choksi, V. (2016). Facial Expression Recognition with Convolutional Neural Networks. \emph{CS231n Course Projects.}

\bibitem{SFEW}
Abhinav Dhall, Roland Goecke, Simon Lucey, and Tom Gedeon. (2011). Static Facial Expressions in Tough Conditions: Data, Evaluation Protocol And Benchmark, \emph{First IEEE International Workshop on Benchmarking Facial Image Analysis Technologies BeFIT, IEEE International Conference on Computer Vision} ICCV2011, Barcelona, Spain, 6-13 November 2011	

\bibitem{Kahou}
Ebrahimi Kahou, S., Michalski, V., Konda, K., Memisevic, R., and Pal, C.  (2015). Recurrent neural networks for emotion recognition in video. \emph{Proceedings of the 2015 ACM on International Conference on Multimodal Interaction}. 467-474. ACM.

\bibitem{AFEW}
Abhinav Dhall, Roland Goecke, Simon Lucey, Tom Gedeon. (2012). Collecting Large, Richly Annotated Facial-Expression Databases from Movies, \emph{IEEE Multimedia}, 19(3):34–41, July 2012

\bibitem{IRNN}
Q. V. Le, N. Jaitly, and G. E. Hinton. (2015). A simple way to initialize recurrent networks of rectified linear units. \emph{arXiv preprint arXiv:1504.00941}.

\bibitem{Ali}
A. Mollahosseini, D. Chan and M. H. Mahoor. (2016).  Going deeper in facial expression recognition using deep neural networks. \emph{2016 IEEE Winter Conference on Applications of Computer Vision (WACV)}, Lake Placid, NY, 1-10.

\bibitem{Ming}
Li M.,  Xu H., Huang X., Song Z., Liu X. and Li X. (2018). \emph{Facial Expression Recognition with Identity and Emotion Joint Learning. IEEE Transactions on Affective Computing.} 1-1.

\bibitem{Tan}
Tan L., Zhang K., Wang K., Zeng X., Peng X. and Qiao Y. (2017) Group emotion recognition with individual facial emotion CNNs and global image based CNNs. \emph{Proceedings of the 19th ACM International Conference on Multimodal Interaction - ICMI 2017}, 549-552. ACM.

\bibitem{AlexNet}
A. Krizhevsky, I. Sutskever, and G. E. Hinton. (2012). Imagenet classification with deep convolutional neural networks. \emph{Advances in neural information processing systems}, 1097–1105.

\bibitem{CKPlus}
Lucey, P., Cohn, J. F., Kanade, T., Saragih, J., Ambadar, Z., \& Matthews, I. (2010). The Extended Cohn-Kanade Dataset (CK+): A complete expression dataset for action unit and emotion-specified expression. \emph{Proceedings of the Third International Workshop on CVPR for Human Communicative Behavior Analysis} (CVPR4HB 2010), San Francisco, USA, 94-101.

\bibitem{Sang}
D. V. Sang, N. Van Dat and D. P. Thuan, "Facial expression recognition using deep convolutional neural networks," 2017 9th International Conference on Knowledge and Systems Engineering (KSE), Hue, 2017, pp. 130-135.

\bibitem{Haar}
Open Source Computer Vision. Face Detection using Haar Cascades: https://docs.opencv.org/3.4.1/d7/d8b/tutorial\_py\_face\_detection.html

\bibitem{Keras}
Chollet, Fran\c{c}ois and others (2015). Keras: https://keras.io/

\bibitem{SKLearn}
Pedregosa et al. (2011). Scikit-learn: Machine Learning in Python, \emph{JMLR 12}, 2825-2830.
\end{thebibliography}
\end{document}